\documentclass[10pt,journal,a4paper]{IEEEtran}
\usepackage{cite}
\usepackage{amsmath,amssymb,amsfonts}
\usepackage{algorithmic}
\usepackage{tikz}
\usepackage{pgfplots}
\usepackage{graphicx}
\usepackage{tabularx}
\usepackage{multirow}
\usepackage{subcaption}
\usepackage{placeins}
\usepackage{hyperref}
\usepackage{url}
\usepackage{textcomp}
\usepackage{xcolor}
\usepackage{pgf}
\usepackage{lmodern}
\usepackage{import}
\usepackage{float}
\def\BibTeX{{\rm B\kern-.05em{\sc i\kern-.025em b}\kern-.08em
    T\kern-.1667em\lower.7ex\hbox{E}\kern-.125emX}}
\begin{document}

\title{Cutup and Detect: Human Fall Detection on Cutup Untrimmed Videos Using a Large Foundational Video Understanding Model}
\author{\IEEEauthorblockN{Till Grutschus\textsuperscript{1}, Ola Karrar\textsuperscript{2}, Emir Esenov\textsuperscript{2} and Ekta Vats\textsuperscript{2}\\}
    \IEEEauthorblockA{\textit{\textsuperscript{1}School of Computation, Information, and Technology, Technical University of Munich, Germany \\ \textsuperscript{2} Department of Information Technology, Uppsala University, Sweden}}
}

\maketitle

\begin{abstract}
    This work explores the performance of a large video understanding foundation model on the downstream task of human fall detection on untrimmed video and leverages a pretrained vision transformer for multi-class action detection, with classes: ``Fall'', ``Lying'' and ``Other/Activities of daily living (ADL)''.
    A method for temporal action localization that relies on a simple cutup of untrimmed videos is demonstrated.
    The methodology includes a preprocessing pipeline that converts datasets with timestamp action annotations into labeled datasets of short action clips.
    Simple and effective clip-sampling strategies are introduced.
    The effectiveness of the proposed method has been empirically evaluated on the publicly available High-Quality Fall Simulation Dataset (HQFSD).
    The experimental results validate the performance of the proposed pipeline.
    The results are promising for real-time application, and the falls are detected on video level with a state-of-the-art 0.96 F1 score on the HQFSD dataset under the given experimental settings.
    The source code will be made available on GitHub.
\end{abstract}


\section{Introduction}
Human fall is an important risk to the health of the elderly, disabled people, and people rehabilitating from injury.
To facilitate support for humans at peril, the field of human fall detection has gained a lot of research interest in recent years \cite{gutierrez_2021, alam_2022}.
The goal is to create a system that can autonomously detect whether a human has fallen and needs help.
To this end, systems employ different data sources, such as sensor data from wearable devices, ambient sensors such as acoustic sensors, or video data from cameras.
With computer vision and video understanding advances, vision-based systems for human fall detection have gained attention \cite{gutierrez_2021}.

\begin{figure*}[!t]
    \centering
    \includegraphics[width=5.5in]
    {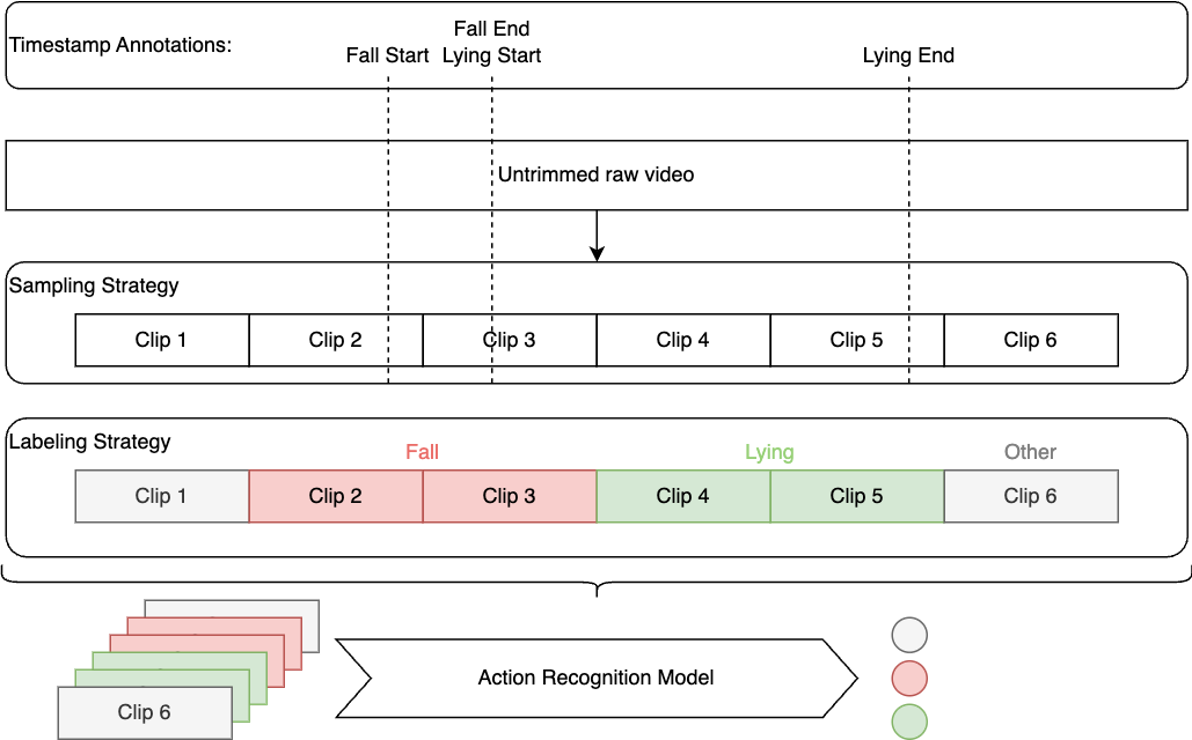}
    \caption{Cutup and detect method pipeline. From the raw untrimmed footage annotated with action timestamps, clips are extracted and labeled with sampling and labeling strategies. The resulting dataset of clips is used in the downstream action recognition model.}
    \label{fig:method_overview}
\end{figure*}

Many works in the field reported good results using hand-crafted features extracted from videos that are subsequently used with traditional classifiers \cite{gutierrez_2021}.
In parallel, research on video understanding in general has shifted towards deep learning-based approaches \cite{hu_2022}.
Unsurprisingly, deep learning-based methods have also gained traction in vision-based human fall detection systems \cite{alam_2022}.

A common approach is experimenting with new architectures and specialized methods \cite{fan_2017, li_2020, alam_2022}.
In other fields, especially natural language processing, a trend has emerged towards using pre-trained large foundational models that are subsequently used for a diverse range of downstream tasks \cite{zhou_2023}.
In this work, we explore the usage of a large foundational model in video understanding for the downstream task of human fall detection instead of applying a specialized architecture.

An essential consideration for vision-based human fall detection systems is their ability to work on untrimmed video streams, such as CCTV footage \cite{li_2020, fan_2017}.
In the literature on video understanding, a well-known preprocessing operation for untrimmed video is temporal action proposal generation (TAPG), i.e., generating candidate clips that are subsequently classified \cite{sooksatra_2022}.
The task of action recognition has shown significant advancements in recent years. However, TAPG still faces critical challenges such as high computational costs and problems with high variability in action classes and action duration \cite{sooksatra_2022, vahdani_2023}.
This work thus focuses on human fall detection on untrimmed videos with naive candidate clip generation, effectively skipping the TAPG preprocessing step.
This has shown promising results in the past \cite{li_2020}.

To that end, we use the public High-Quality Fall Simulation Dataset (HQFSD) \cite{baldewijns_2016} with augmented annotations to train a large foundational video understanding model VideoMAEv2 - a pretrained vision transformer \cite{wang_2023} to detect three action classes: ``Fall'', ``Lying'' and ``Other/Activities of daily living (ADL)''.
During training, we test two simple clip-sampling strategies: Cutup and Gaussian sampling.

The main contributions of this work are as follows:
\begin{itemize}
    \item Our work advances the state-of-the-art in human fall detection on untrimmed video data from HQFSD, which is selected in this study as a challenging dataset that enables generalization to real-world applications.
    \item We propose and demonstrate a method for temporal action localization that relies on a simple cutup of untrimmed videos.
    \item We present a preprocessing pipeline that converts datasets with timestamp action annotations into labeled datasets of short action clips.
    \item Our proposed methodology explores the performance of a large video understanding foundation model on the downstream task of human fall detection and demonstrates that it can outperform previous specialized architectures.
\end{itemize}

\section{Related works}

In the literature, fall detection belongs to the broader category of human action detection.
The following action detection tasks can be distinguished \cite{hu_2022}:
\begin{itemize}
    \item Action Detection (AD): Classification of \textit{trimmed videos} into action categories. Trimmed videos are pre-segmented clips containing only a single action instance \cite{vahdani_2023}.
    \item Temporal Action Detection (TAD): Besides classifying action categories, the action's start and end time must be found in an \textit{untrimmed video}. In the context of human fall detection, this involves detecting the start and end of a fall in a long video.
    \item Spatial-Temporal Action Detection (STAD): In addition to identifying the start and end of an action, this task involves locating the spatial position of the action in each video frame. In the case of falling, this would entail finding a bounding box around the person falling throughout all frames in the video \cite{hu_2022}.
\end{itemize}
This work focuses on temporal action detection of falls in \textit{untrimmed video}.

Deep learning-based methods are gaining much attention in this broader field because of their remarkable performance and effectiveness in extracting features from multi-dimensional datasets \cite{morshed_2023}.
These methods employ large foundation encoder models that learn to generate abstract representations for action recognition and have become dominant over the hand-crafted feature extraction approach prevalent in the past \cite{bobick_2001, laptev_2003, heng_2011, heng_2013}.
These methods can be categorized into two-stream Convolutional Neural Networks (CNN) \cite{simonyan_2014, peng_2017, wang_2019}, Recurrent Neural Networks (RNN) \cite{sun_2017, ullah_2018, he_2020, li_2016, meng_2019}, 3D-CNN \cite{tran_2014, tran_2017, carreira_2017, hara_2017, feichtenhofer_2020}, and Transformer-based methods \cite{dosovitskiy_2020, arnab_2021, bertasius_2021, vaswani_2017, yan_2022, wang_2023}.
In this work, we use a large transformer-based model.

For human fall detection, previous works mainly focus on specialized architectures.
In \cite{fan_2017},  dynamic images are calculated from sub-videos generated with a sliding window and fed into a CNN.
The work \cite{li_2020} implements a similar approach where dynamic pose motion representations are used and fed into a CNN-LSTM.
Both works perform well by simply using a sliding window over the untrimmed video.
This sliding window approach effectively turns the TAD task into many AD tasks. This paper uses a similar approach, i.e., \textit{Cutup sampling}.
Instead of devising a specialized architecture for human fall detection, we leverage one of the previously described large foundational models for this task.

Several vision-based human fall detection datasets have been published \cite{alam_2022}.
Datasets commonly used as benchmarks include the Le2i Fall Detection Dataset (Le2i FDD), the University of Rzeszow Fall Detection (URFD) Dataset, the Multiple Cameras Fall Dataset (MCFD), and the High-quality fall simulation data (HQFSD) \cite{li_2020, fan_2017, alam_2022}.
The HQFSD stands out as a challenging dataset, including multi-person scenarios, changing lighting, occlusion, and simulating a real-world environment \cite{baldewijns_2016, alam_2022}.
As presented in \cite{fan_2017, li_2020, debard_2015}, the existing human action recognition models perform well for other benchmark datasets but struggle to achieve good performance on HQFSD.
Given its challenges and comparably poor performance in previous work, we select the HQFSD as a benchmark in this work.

\begin{figure*}[!t]
    \centering
    \includegraphics[width=6.5in]{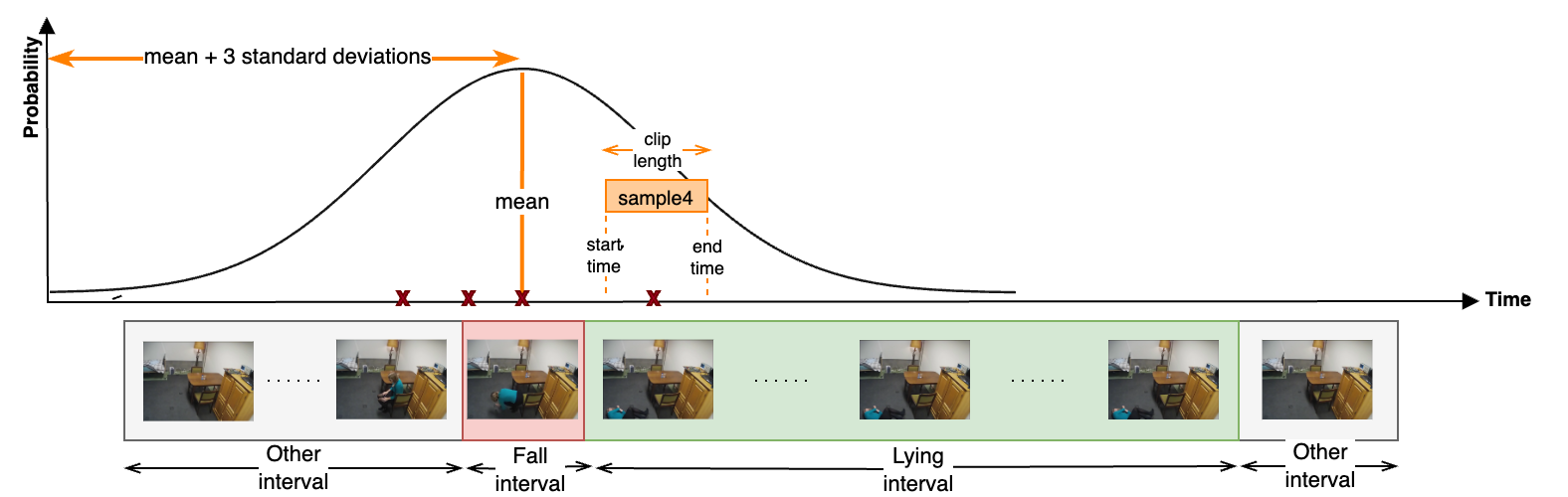}
    \caption{Illustration of Gaussian sampling for a single video within the HQFSD. The red crosses illustrate sampled seeds $t_i$. Figure best viewed in colors.}
    \label{fig:GaussianSampling}
\end{figure*}

\section{Method}
The overall pipeline of our proposed cutup and detect method is presented in Figure \ref{fig:method_overview}. From the raw untrimmed video footage as an input, annotated with action timestamps, clips are extracted and labeled using sampling and labeling strategies. Specifically, we introduce Cutup sampling and Gaussian sampling strategies. The resulting dataset of clips is then used in the downstream action recognition model.

The following sections present the individual components: sampling strategies, labeling strategy, and the action recognition model.

\subsection{Clip Sampling Strategy} \label{sec:sampling-strategy}
To extract short clips from the untrimmed videos in the dataset, we sample equal-length clips from each raw video.
Random undersampling of specific classes is supported for all types of sampling strategies.
The specifics of the two sampling strategies are outlined below:

\paragraph{Cutup Sampling}
During Cutup sampling, the video is cut into equal-length clips by a sliding window, similar to previous works \cite{li_2020, fan_2017}.
Hyperparameters define the clip length, overlap, and stride.

\paragraph{Gaussian Sampling}
Additionally, we introduce Gaussian sampling. Given an untrimmed fall video $V$ of length $T$ seconds, in which the middle of the falling interval is at the second  $t_{Fall}$, we sample $n =\left \lceil{\frac{T}{T_{clip}}} \right \rceil$ clips of length $T_{clip}$.
Specifically, we sample $n$ seeds from  a Gaussian distribution as follows:
\begin{equation*}
    t_i \sim \mathcal{N}\left(t_{Fall} , \frac{1}{3}\min \{t_{Fall}, (T-t_{Fall})\}\right) \, \text{for}\, i = \{1, \dots, n\}
\end{equation*}

\noindent where the mean is $t_{Fall}$, and the standard deviation is one-third of the minimum between the time from the start of the video to the middle of the fall interval and the time from the middle of the fall interval to the end of the video.
The seeds are then used as the middle points of the clips
\begin{equation*}
    \operatorname{c}_i = (t_i -\frac{1}{2}T_{clip}, t_i +\frac{1}{2}T_{clip})
\end{equation*}

This way, the samples cover most of the raw video.
For the case of a video without a fall, a backup sampler must be specified. We use Cutup sampling as the backup sampler.
Figure \ref{fig:GaussianSampling} demonstrates the application of Gaussian sampling for a single video within the HQFSD.

Since Gaussian sampling requires knowledge of the annotation, it must only be used during training set generation.

\subsection{Labeling Strategy}
After sampling clips from the video, each clip is assigned a label.
We adopt a priority labeling approach, where the fall action has the highest priority, followed by lying, and finally, ADL or other activities.
For instance, if a part of the clip contains falling and lying or other activities, the clip will be labeled as falling.
Similarly, if a clip contains both lying and any ADL or other activities, excluding falling, it will be labeled as lying.

\subsection{Action Recognition Model} \label{sec:action-recognition-model}
\begin{figure}[!t]
    \centerline{\includegraphics[width=\linewidth]{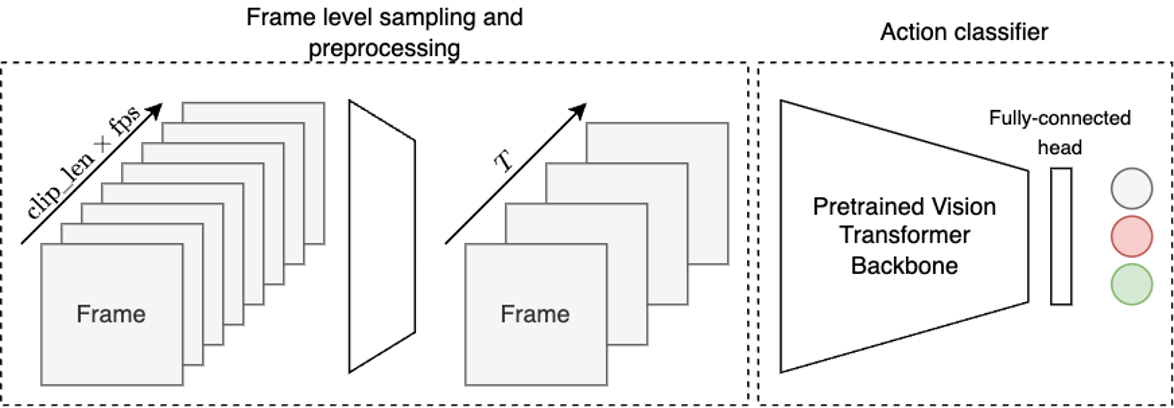}}
    \caption{Overview of the action recognition pipeline. The clips are preprocessed, and individual frames are sampled. A vision transformer backbone is used for feature extraction, and a fully connected head is used for classification.}
    \label{fig:model_overview}
\end{figure}

The complete pipeline for classifying individual clips retrieved from the sampling process is presented in Figure \ref{fig:model_overview} and is discussed as follows.

\paragraph{Frame-level sampling and preprocessing}
Before extracting embeddings with the VideoMAEv2 backbone, the individual clips are preprocessed, and individual frames are sampled.
From the total of $\text{clip\_len} \times \text{fps}$ frames available in a clip, 16 frames are sampled as required by the pre-trained backbone \cite{wang_2023}.
This sampling operation is parametrized by the number of frames per sample $N=16$, the sampling stride $\tau$, and the number of samples per clip $C$.
The number of samples per clip $C$ is set to $1$ during training and validation.
At test time, $C=5$ samples are taken from each clip.
During training, the frames are resized to $W \times 224$ pixels and random-cropped to $224 \times 224$ pixels. Additionally, flip augmentation with $p=0.5$ is applied to each sample.
For reproducibility, validation samples are not flipped and center-cropped to size.
During inference, samples are not flipped; instead of center-cropping, a three-crop operation is applied.
Three horizontal crops of size $224\times 224$ are taken from the resized $W \times 224$ frames.
By taking $C=5$ samples per clip and three crops of each sample, we effectively run $15$ predictions per test clip.
The scores are generated by averaging the scores of each prediction.
The described process is widely used in the literature \cite{wang_2023, mmaction2_2020, bertasius_2021}.

Finally, we have calculated the training dataset's pixel-wise first and second statistical moments \cite{hancock_2017} to normalize the input frames.

\paragraph{VideoMAEv2 Vision Transformer backbone}
To generate embeddings from the preprocessed frames, a VideoMAEv2 model is used \cite{wang_2023}.
This is a large video understanding foundation model that has been pre-trained on a large-scale video dataset in a self-supervised manner.
It has shown state-of-the-art performance in different benchmarks and tasks \cite{wang_2023}.
The underlying architecture is a vision transformer \cite{dosovitskiy_2020}.
We use a distilled version (ViT-B) with 87M parameters.
The weights used result from a fine-tuned model for action recognition on the Kinetics400 benchmark \cite{mmaction2_2020}.

\paragraph{Classification head}
A single fully connected layer classifies the embeddings.
We use a standard cross-entropy loss.
As described in the frame-level sampling description above, the classification head averages the logits of multiple clips during inference.

The pipeline and model have been implemented in the MMAction2 framework \cite{mmaction2_2020}.

\section{Experimental Results}
\label{sec:experiments}
This section describes the dataset and annotation procedures, the overall experimental setup, and demonstrates the performance of the proposed method on HQFSD.

\subsection{Dataset}

\paragraph{High-quality fall simulation Data}
``High-quality fall simulation data'' (HQFSD) \cite{baldewijns_2016} is an open-source dataset containing 2:25:54 hours of fall data from 55 different fall scenarios and 5:50:29 hours of Activities of daily living (ADL). Each fall scenario has an average length of 2:45 min.
The dataset was created to address several shortcomings of existing human fall datasets, which often fail to include difficulties with visual monitoring systems deployed in real settings.
Difficulties addressed include changes in illumination and occlusions, video length, and environmental challenges such as furniture and other objects in the video.
The data was collected from five web cameras recording a single room.
The fall scenarios include different walking aids, fall speeds, moving objects during the fall, and starting- and stopping poses.

We chose the dataset due to its quality and the many challenges included in the data.
While less realistic datasets can result in good benchmark results in human action recognition models, these might fail to generalize well in more natural settings.

\paragraph{Annotation}
The HQFSD is annotated with the start and end times of the fall.
However, additional annotations are required.
Given that our action recognition labels include ``Falling'', ``Lying'', and ``ADL / Other'', we require the timestamp for the end of the lying phase.
The end of the lying phase is the moment when the actor starts getting back up at the end of every fall scenario.
We reviewed every scenario and annotated the second in which the actor starts to get up again.
For scenarios in which the actor crawls after the fall, we annotated the second the actor starts crawling.

In addition, not every fall or lying action is visible from all camera angles.
The authors of the HQFSD have ranked the different camera angles for every scenario based on how well the actors are visible.
To avoid annotating videos as a fall, even if nothing can be seen on the video, we added two flags to every video: ``Fall visible'' and ``Lying visible''.
We set these flags by first checking the camera angle with the lowest ranking.
If everything is visible from the worst angle, we assume that both falling and lying are visible from all other cameras.
Conversely, if something cannot be seen from the lowest-ranking camera angle, we additionally checked the next higher-ranking camera angle.
Camera angles with only partial visibility of the lying body are still considered valid.

\subsection{Experimental Setup}

The videos are split into training ($70\%)$, validation ($20\%$), and testing ($10\%$) subsets.
Since ADL videos are much longer, we stratified the split on the video type, i.e., fall videos and ADL videos.
This ensures that the total length of videos in a dataset roughly reflects the intended split.
Consequently, every dataset has approximately $24\%$ ADL videos and $76\%$ fall videos.

The fine-tuning is run on a single Nvidia A40 GPU.
The training hyperparameters are mostly taken from \cite{wang_2023}; an overview can be found in Table \ref{tab:training-hyperparams}.

To find parameters for the preprocessing pipeline, we ran five experiments and tested performance with a faster cosine decay of 35 epochs.
Appendix \ref{apx:additional-experiments} provides details on these experiments and results. The parameters used in the final evaluation of our method are presented in Table \ref{tab:overview-experiments}.

\begin{table}[!t]
    \caption{Hyperparameters used for fine-tuning.}
    \begin{center}
        \begin{tabularx}{\linewidth}{l|X}
            \hline
            \textbf{Config}             & \textbf{Value}                                                     \\
            \hline
            optimizer                   & AdamW                                                              \\
            base learning rate          & 7e-4                                                               \\
            weight decay                & $0.05$                                                             \\
            optimizer momentum          & $\beta_1, \beta_2 = 0.9, 0.999$                                    \\
            batch size$^{\mathrm{a}}$   & $12$                                                               \\
            learning rate schedule      & cosine decay                                                       \\
            warmup epoch$^{\mathrm{b}}$ & $5$                                                                \\
            max epochs                  & $90$                                                               \\
            drop path                   & $0.3$                                                              \\
            layer-wise lr decay         & 0.75                                                               \\
            \hline
            \multicolumn{2}{l}{$^{\mathrm{a}}$ Limited by GPU memory; Scaled base learning rate accordingly} \\
            \multicolumn{2}{l}{$^{\mathrm{b}}$ Linear interpolation of scaling factor from 1e-3 to 1}        \\
        \end{tabularx}
        \label{tab:training-hyperparams}
    \end{center}
\end{table}

\begin{table}[t]
    \caption{Overview of Cutup and Detect parameters.}
    \begin{center}
        \begin{tabularx}{\linewidth}{l|>{\centering\arraybackslash}X>{\centering\arraybackslash}X}
            \hline
            \textbf{Config}              & \multicolumn{2}{c}{\textbf{Experiment}}                     \\
            \hline
            Sampling strategy            & {Cutup}                                 & {Gaussian}        \\
            \hline
            Fallback sampler             & -                                       & {Cutup, stride 5} \\
            \hline
            Clip length                  & \multicolumn{2}{c}{5s}                                      \\
            \hline
            Random undersampling         & \multicolumn{2}{c}{30\% fall class}                         \\
            \hline
            Frame sampling stride $\tau$ & \multicolumn{2}{c}{8}                                       \\
            \hline
        \end{tabularx}
        \label{tab:overview-experiments}
    \end{center}
\end{table}

\subsection{Results}
We report the results of the two models on the test dataset for the clip level and aggregated to the video level for comparison with other works.
All tests were run on an Nvidia Tesla T4 GPU.
The average inference time for a 5s clip was 2.46s, making the system capable of real-time application.

\paragraph{Clip-level classification}
We first evaluated the model on the test dataset. We used Cutup sampling with a clip length of 5 to generate the test samples.
Table \ref{tab:clip-level-test-performance} highlights the achieved performance.

Notably, the model reliably predicts clips at over $0.9$ average F1 for both sampling methods.
The recall of $0.82$ for Gaussian sampling is especially promising since, in a production scenario, alarms may only be triggered when a fall has been detected, and subsequent clips contain lying. Therefore, a high recall is preferable over a high precision.

There is only a small difference between the two models.
Generally, Cutup sampling leads to an overall better performance with significantly better precision in detecting falls.
Gaussian sampling improves the recall.
This result is unsurprising, given that the model trained with a Gaussian sampled dataset has seen a more extensive variety of fall clips.

Nevertheless, we hypothesized that Gaussian sampling would outperform Cutup sampling in both metrics.
Possibly, this is not the case because the Gaussian sampling process does not sample from the full video and thus does not represent the input video in its full variety.
If the fall occurs in the first half of the raw video, there is a very low probability for clips sampled from the end of the video. Since this is typically where the actors stand up again, the Gaussian sampling strategy may not sufficiently capture this action.
This result shows the importance of having a wide variety of actions in the training data.

\begin{figure*}[!t]
    \centering
    \begin{subfigure}{\textwidth}
        \centering
        \includegraphics[width=.24\linewidth]{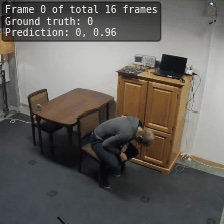}
        \includegraphics[width=.24\linewidth]{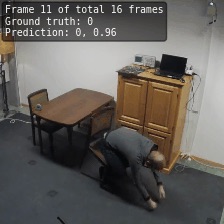}
        \includegraphics[width=.24\linewidth]{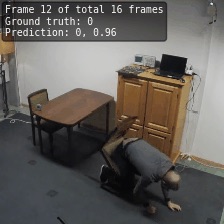}
        \includegraphics[width=.24\linewidth]{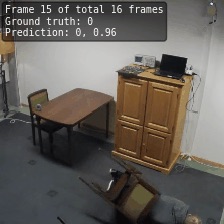}
        \caption{Fall}
    \end{subfigure}
    \vspace{0.1cm}

    \begin{subfigure}{\textwidth}
        \centering
        \includegraphics[width=.24\linewidth]{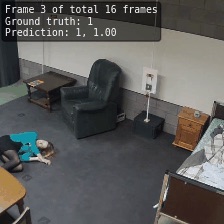}
        \includegraphics[width=.24\linewidth]{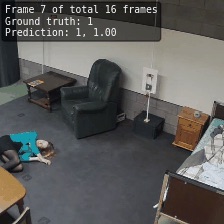}
        \includegraphics[width=.24\linewidth]{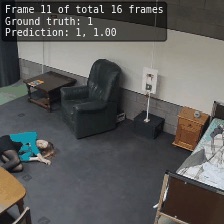}
        \includegraphics[width=.24\linewidth]{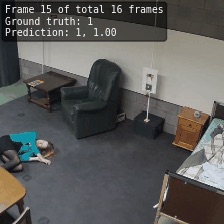}
        \caption{Lying}
    \end{subfigure}
    \vspace{0.1cm}

    \begin{subfigure}{\textwidth}
        \centering
        \includegraphics[width=.24\linewidth]{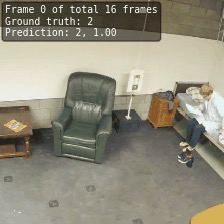}
        \includegraphics[width=.24\linewidth]{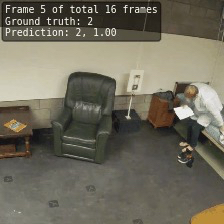}
        \includegraphics[width=.24\linewidth]{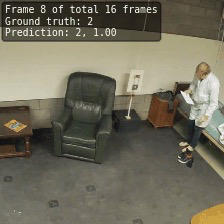}
        \includegraphics[width=.24\linewidth]{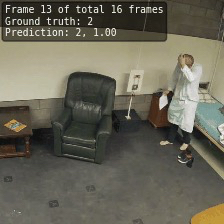}
        \caption{ADL}
    \end{subfigure}
    \caption{Visualisation of results for selected frames from different clips.}
    \label{fig:example-samples}
\end{figure*}

Figure \ref{fig:example-samples} presents the visualization of results for selected frames from different clips for the three categories - fall (label 0), lying (label 1), and ADL (label 2), along with the class logits.

\paragraph{Video-level classification}
To gauge our model's performance in the video-level classification task and compare our method to other works, we aggregated the model's predictions to the video level.
Note, however, that our method is inherently not a binary classification approach.
A real-world application of our method can additionally indicate whether a person is not moving after a fall and needs help.

The aggregation is done as follows. If any clip from a video is classified as a fall, we mark the whole video as a fall video. Otherwise, i.e., if only ``Lying'' or ``Other'' clips are predicted, we mark the whole video as ``Activities of daily living (ADL)''.

\begin{table}[!t]
    \caption{Performance of the model using Gaussian Sampling and Cutup Sampling during training on the test dataset.}
    \begin{center}
        \begin{tabularx}{\linewidth}{X|X|cccc}
            \hline
            \multicolumn{2}{l}{ }     & Precision          & Recall        & F1-Score      & Support              \\
            \hline
            \multirow{4}{*}{Cutup}    & Fall               & \textbf{0.94} & 0.78          & \textbf{0.85} & 74   \\
                                      & Lying              & 0.99          & \textbf{0.97} & \textbf{0.98} & 983  \\
                                      & Other              & 0.99          & 1.00          & 0.99          & 4544 \\ \cline{2-6}
                                      & Avg$^{\mathrm{a}}$ & \textbf{0.97} & 0.92          & \textbf{0.94} & 5601 \\
            \hline
            \multirow{4}{*}{Gaussian} & Fall               & 0.73          & \textbf{0.82} & 0.78          & 74   \\
                                      & Lying              & 0.99          & 0.96          & 0.97          & 983  \\
                                      & Other              & 0.99          & 1.00          & 0.99          & 4544 \\ \cline{2-6}
                                      & Avg$^{\mathrm{a}}$ & 0.91          & \textbf{0.93} & 0.91          & 5601 \\
            \hline

            \multicolumn{6}{l}{$^{\mathrm{a}}$unweighted}
        \end{tabularx}
        \label{tab:clip-level-test-performance}
    \end{center}
\end{table}

\begin{table}[!t]
    \caption{Fall detection performance aggregated to the video level and comparison with other works base on the reported precision (Prec), sensitivity(Sens), specificity (Spec) and F$1$ score (F$1$)  }
    \begin{center}
        \begin{tabularx}{\linewidth}{X|cccc}
            \hline
                                             & Prec          & Recall/Sens   & Spec          & F1            \\
            \hline
            Debard et al. \cite{debard_2015} & -             & 0.62          & 0.41          & 0.60          \\
            Fan et al. \cite{fan_2017}       & -             & 0.74          & 0.69          & -             \\
            Li et al. \cite{li_2020}         & 0.81          & 0.96          & -             & 0.88          \\
            \hline
            Cutup                            & \textbf{0.96} & 0.93          & \textbf{0.88} & 0.94          \\
            Gaussian                         & 0.95          & \textbf{0.98} & 0.82          & \textbf{0.96} \\
            \hline
        \end{tabularx}
        \label{tab:performance-comparison}
    \end{center}
\end{table}

Table \ref{tab:performance-comparison} presents our method's resulting video-level fall detection performance compared with other works.
Our method outperforms the previous state-of-the-art on the HQFSD in both precision and recall using the Gaussian sampling strategy.

\section{Conclusion}
This work successfully implemented a large foundation video understanding model for human fall detection on untrimmed video data.
The experimental results demonstrate the effectiveness of the proposed method on the publicly available High-quality fall simulation data (HQFSD) and indicate the potential for real-time application on CCTV footage.
To the best of the authors' knowledge, we outperform the current state-of-the-art for video-level classification on the HQFSD by achieving \textbf{0.96} F1 score under the given experimental settings.
The source code and pre-trained models will be made available for the research community on GitHub.

\textbf{Limitations and Future Work:} Given that the train and test splits were generated based on the videos themselves and not the individual scenarios in the dataset, some information from the training data may have spilled over to the test data since certain videos show the same actions from different angles.
This may have induced unwanted dependence on the data, which in turn may have inflated our test scores.
Data splits where no scenarios from the training data are replicated in the test data, even if shot from different angles, would guarantee the absence of this unwanted effect.
Other works appear to have the same limitation \cite{li_2020}. Splitting the dataset by scenario would yield a small training set, making training more difficult.

Additionally, our method addresses the problem of temporal action detection.
However, the temporal resolution is limited by the clip length and stride. Lastly, experiments on other datasets are left for future work.

\section*{Acknowledgment}
This research was supported by \emph{Kjell och M{\"a}rta Beijer Foundation}.
The computations were enabled by resources provided by the National Academic Infrastructure
for Supercomputing in Sweden (NAISS) partially funded by the Swedish Research Council through
grant agreement no. 2022-06725.

\FloatBarrier
\appendices

\section{Additional Experiments}
\label{apx:additional-experiments}

We additionally report training results on a total of 5 experiments. These experiments were used to find suitable parameters for our cutup and detect pipeline.
Table \ref{tab:overview-additional-experiments} gives an overview of the experiments.
The class weighting for the cross-entropy loss in experiment 2 is calculated as follows.
Given the total number of samples in training set $N$ and the number of samples for class $i$ $n_i$, the weight for class $i$ is given by $N/(n_i * c)$ with $c=3$ classes.
Random undersampling is done for the ``Fall'' class.

Parameters ``Sampling Strategy'', ``Fallback Sampler'', ``Clip length'', and ``Random undersampling'' do not directly influence the model or individual samples' preprocessing.
Instead, they control how the clips are generated from the untrimmed video.
Hence, we provide a ``Data code'' to differentiate the generated training sets.

In all experiments with Gaussian sampling, the validation dataset was generated using Cutup sampling with identical clip lengths.

\begin{table}[t]
    \caption{Overview of five experiments conducted with different configurations.}
    \begin{center}
        \begin{tabularx}{\linewidth}{l|>{\centering\arraybackslash}X|>{\centering\arraybackslash}X|>{\centering\arraybackslash}X|>{\centering\arraybackslash}X|>{\centering\arraybackslash}X}
            \hline
            \textbf{Config}              & \multicolumn{5}{c}{\textbf{Experiment Number}}                                                                                               \\
            \cline{2-6}
                                         & \textbf{1.}                                    & \textbf{2.}                            & \textbf{3.}            & \textbf{4.} & \textbf{5.} \\
            \hline
            Sampling strategy            & \multicolumn{2}{c|}{Cutup}                     & \multicolumn{3}{c}{Gaussian}                                                                \\
            \hline
            Loss weighting               & No                                             & Yes                                    & \multicolumn{3}{c}{No}                             \\
            \hline
            Fallback sampler             & \multicolumn{2}{c|}{-}                         & \multicolumn{3}{c}{Cutup$^\mathrm{a}$}                                                      \\
            \hline
            Clip length                  & \multicolumn{3}{c|}{10s}                       & \multicolumn{2}{c}{5s}                                                                      \\
            \hline
            Random undersampling         & \multicolumn{3}{c|}{-}                         & \multicolumn{2}{c}{30\%}                                                                    \\
            \hline
            Frame sampling stride $\tau$ & \multicolumn{4}{c|}{4}                         & 8                                                                                           \\
            \hline
            Data code                    & \multicolumn{2}{c|}{A}                         & B                                      & \multicolumn{2}{c}{C}                              \\
            \hline
            \multicolumn{6}{l}{$^\mathrm{a}$for 5s clip length, stride is set to 5}                                                                                                     \\
        \end{tabularx}
        \label{tab:overview-additional-experiments}
    \end{center}
\end{table}
\begin{table}[!t]
    \caption{Label distributions of the training datasets generated by the different sampling pipelines.}
    \begin{center}
        \begin{tabularx}{\linewidth}{l|rrr}
            \hline
            Data code: & A              & B              & C               \\
            Class      &                &                &                 \\
            \hline
            Fall       & 231 (2.2\%)    & 1,221 (11.3\%) & 1,590 (11.4\%)  \\
            Lying      & 1,769 (16.4\%) & 835 (7.8\%)    & 2,085 (14.9\%)  \\
            Other      & 8,764 (81.4\%) & 8,708 (80.9\%) & 10,296 (73.7\%) \\
            \hline
            Total      & 10,764 (100\%) & 10,764 (100\%) & 13,971 (100\%)  \\
            \hline
        \end{tabularx}
        \label{tab:label-distributions}
    \end{center}
\end{table}

\subsection{Dataset Analysis}
Given that our pipeline mainly influences the way training and test data is generated, we give some insight into the resulting datasets.

\paragraph{Label distribution}
Table \ref{tab:label-distributions} shows the label distributions of the datasets of clips. The codes A, B, and C result from the different experimental setups described in Table \ref{tab:overview-additional-experiments}.

Dataset A shows a severe class imbalance.
Cutup sampling only generates 231 ``Fall'' samples from the untrimmed videos in the training set.
Gaussian sampling (B and C) severely reduces the class imbalance.
Additionally, a visual investigation of individual samples reveals that the fall occurs in more diverse positions throughout the clips.
Finally, reducing the clip length (C) increases the dataset size and thus increases class imbalance.
This is compensated by random undersampling of the ``Other'' class.

\paragraph{Label quality}
Given that the labels are automatically assigned according to the labeling strategy, the labels can be imprecise.
While there is no way of quantifying the label quality without visually inspecting every sample, a qualitative inspection of the pipeline output showed some noteworthy results.

Since the model only samples 16 frames from the clips, the correctness of the ``Fall'' label is mainly influenced by the probability that one or more of the 16 sampled frames depict the fall.
Given a sampling stride $\tau$, the window from which the individual frames are sampled has a size of $S = \tau \times 16$ frames.
The total number of frames in a clip (produced by the sampling strategy) is $N = T_{clip} \times \mathrm{fps}$.
The priority labeling strategy assigns the label ``Fall'', even if only a single frame shows the fall.
Hence, the worst-case probability of the model seeing a frame that contains a fall is given by  $S/N$.
Decreasing the clip length and increasing the sampling stride will increase this probability, effectively increasing the label quality.
Other factors, such as a threshold for label assignment in the labeling strategy, can also influence the label quality.

Finally, we can see that reducing the clip length (C) potentially increases class imbalances and label quality. A visual inspection of the pipeline outputs confirmed this.

\subsection{Training and validation performance}
Table \ref{tab:validation-f1-scores} shows the F1 scores on the validation sets for the epochs with the best macro average F1.

\begin{table}[!t]
    \caption{F1 scores on the respective validation sets during training. The corresponding epoch is in brackets.}
    \begin{center}
        \begin{tabularx}{\linewidth}{l|rrrr}
            \hline
            \textbf{Experiment ID} & Fall          & Lying         & Other         & Macro Average \\
            \textbf{(Epoch)}       &               &               &               &               \\
            \hline
            1 (61)                 & 0.63          & 0.90          & \textbf{0.98} & 0.84          \\
            2 (31)                 & 0.45          & 0.91          & 0.97          & 0.78          \\
            3 (9)                  & 0.39          & 0.90          & \textbf{0.98} & 0.76          \\
            4 (11)                 & 0.52          & 0.89          & \textbf{0.98} & 0.80          \\
            5 (16)                 & \textbf{0.74} & \textbf{0.92} & \textbf{0.98} & \textbf{0.88} \\
            \hline
        \end{tabularx}
        \label{tab:validation-f1-scores}
    \end{center}
\end{table}

Experiment 5 shows the best performance.
Interestingly, as initially hypothesized, performance did not increase from experiment 1 to experiment 3.
The difference between experiments 1 and 3 is the switch to Gaussian sampling (data code A to data code B), which decreased class imbalance.
We hypothesized that the class imbalance is the most crucial factor in the preprocessing pipeline.
However, these results suggest otherwise.

In the subsequent experiments, 4 and 5, the performance was recovered.
The difference between experiments 4 and 5 compared to experiment 3 is the reduction of the clip length and increase in frame-level sampling stride, directly increasing the label quality.
This observation indicates that the label quality strongly influences the performance.

\FloatBarrier
\bibliography{refs}
\bibliographystyle{ieeetr}

\end{document}